\documentclass[11pt]{article}
\usepackage[T1]{fontenc}
\usepackage{times} 
\usepackage{mathptmx}
\usepackage{textcomp}

\usepackage{blindtext}

\usepackage{graphicx}%
\usepackage{multirow}%
\usepackage{amsmath,amssymb,amsfonts}%
\usepackage{amsthm}%
\usepackage{mathrsfs}%
\usepackage[title]{appendix}%
\usepackage{xcolor}%
\usepackage{textcomp}%
\usepackage{manyfoot}%
\usepackage{booktabs}%
\usepackage{algorithm}%
\usepackage{algorithmicx}%
\usepackage{algpseudocode}%
\usepackage{listings}%
\usepackage{comment}%
\usepackage{subcaption}

\usepackage{amsmath,amssymb}
\usepackage{bbm}
\usepackage{comment}

\newcommand{\Bc}{\mathcal{B}}

\newcommand{\Eb}{\mathbbm{E}}

\newcommand{\Gc}{\mathcal{G}}

\newcommand{\Pc}{\mathcal{P}}

\newcommand{\Xc}{\mathcal{X}}
\newcommand{\Yc}{\mathcal{Y}}
\newcommand{\Ic}{\mathcal{I}}
\newcommand{\Wc}{\mathcal{W}}
\newcommand{\Zc}{\mathcal{Z}}

\newcommand{\1}{\mathbbm{1}}
\newcommand{\argmax}{\mathop{\rm argmax}}

\newcommand{\D}{\mathop{\text{\rm d}\!}}

    \definecolor{plum}{rgb}{0.3,0,0.7}

\usepackage{lastpage}

\title{Federated Calculation of the Free-Support Transportation Barycenter by Single-Loop Dual Decomposition}

\author{Zhengqi Lin and Andrzej Ruszczy\'nski \footnote{Department of Management Science and Information Systems, Rutgers University, email: zhengqi.lin@rutgers.edu;rusz@rutgers.edu}}

\begin{document}

\maketitle

\begin{abstract}
 We propose an efficient federated dual decomposition algorithm for calculating the Wasserstein barycenter of several distributions, including choosing the support of the solution. The algorithm does not access local data and uses only highly aggregated information.
It also does not require repeated solutions to mass transportation problems.
 Because of the absence of any matrix-vector operations, the algorithm exhibits a very low complexity of each iteration and significant scalability. We illustrate its virtues and compare it to the state-of-the-art methods on several examples of mixture models.
\end{abstract}

\section{Introduction}

The Monge--Kantorovich transportation distance, frequently referred to as the ``Wasserstein'' metric or the ``Earth-Mover's'' distance, is a distance between probability measures on a given space $\Yc$ with metric $d(\cdot,\cdot)$.
For  $p \ge 1$,
the $p$th order distance between two measures $\mu$ and $\nu$ that have finite $p$th moments (belong to the space $\Pc_p(\Yc)$) is defined as:
\begin{equation} \label{Wass}
W_{p}(\mu, \nu) =\left(\min _{\pi \in \Pi(\mu, \nu)} \int_{\mathcal{Y} \times \mathcal{Y} } d(y, z)^{p} \;  \pi({\D y}, {\D z})\right)^{1 / p},
\end{equation}
where $\Pi(\mu, \nu)$ is the set of all probability measures on $\Yc\times\Yc$ with the marginals $\mu$ and $\nu$.
We refer the reader to
\cite{villani2009optimal} for the historical account and comments on the terminology.

In recent years, the transportation distance has demonstrated promising results in various applications, including Generative Adversarial Networks \cite{arjovsky2017wasserstein}, clustering \cite{ho2017multilevel}, semi-supervised learning \cite{pmlr-v32-solomon14}, and image retrieval \cite{Rubner2000TheEM,5459199}. Furthermore, recent contributions have extended its applicability to measuring the distance between mixture distributions, such as the sketched Wasserstein distance proposed by \cite{SW} for finite mixture models, and distances tailored to Gaussian mixture models \cite{GMM1, GMM2, GMM3}.

In this article, we are concerned with a \emph{family} of probability measures on $\Yc$\!, parametrized by an argument $x\in \Xc$, where $\Xc$ is another metric space.
Formally, such a family can be viewed as a \emph{stochastic kernel}: a measurable mapping $Q$ from $\Xc$ to the space $\Pc_p(\Yc)$ of probability measures on $\Yc$
having finite $p$th moments. For
a fixed distribution $\lambda$ on $\Xc$ and any probability measure $q$ on $\Yc$, we can define the $p$th  order distance between the kernel $Q$ and the measure $q$ as follows
\begin{equation} \label{TS}
    \Wc_{p}^\lambda(Q, q)=\left(\int_{\Xc} \big[{W}_{p}(Q(x), q)\big]^p \;{\lambda}(\D x)\right)^{1/p} .
\end{equation}
The function $\Wc_{p}^\lambda(Q, q)$ calculates an ``average'' transportation distance between $Q(x)$ and $q$, over all parameter values $x$, weighted
by the measure $\lambda$. It
 is a special case of the Integrated Transportation Distance between stochastic (Markov) kernels, recently introduced in \cite{lin2023integrated}.
There is also a connection to the weak transport theory of \cite{backhoff2022applications}.

Using the distance \eqref{TS}, we define the $p$th order \emph{transportation barycenter problem}
as follows:
\begin{equation}
\label{barycenter}
\min_{q\in \Pc_p(\Yc)} \Wc_{p}^\lambda(Q, q).
\end{equation}
It is a generalization of the Wasserstein barycenter problem introduced in \cite{Agueh2011} for discrete distributions $\lambda$ and $p=2$.

{

\subsubsection*{Business and Data Analysis Applications: Fairness}

Our interest in the barycenter problem is due to its relevance for manifold business applications. In finance and insurance, the parameter $x$ may represent a particular location,
affiliated business, or a specific customer group, and $Q(x)$ may represent the distribution of a certain characteristic of customers with this designation.
Then, the barycenter $q$ may serve as
a representative distribution for policy development.

For example, let $X\in\Xc$ represent a sensitive information of an insured, such as race or gender, $Z$ the other risk factors, and $L$ the loss. For simplicity, we
assume that $\Xc$ is finite. The pure premium is
defined as the \emph{Bayes predictor} $\mu(X,Z) = \Eb[L|X,Z]$. In general, the conditional distribution of $\mu(X,Z)$, given $X=x$, is different for different $x$, which may be undesirable or outright illegal.
Regressors $g(X,Z)$ having the property that $\Eb[g(X,Z)|X=x]$ is the same for all $x$ are called \emph{fair} (see, \cite{charpentier2023mitigating,carassai2024neural}).
We denote their class as $\Gc^{\text{fair}}$. The
calculation of the fair premium can be now formulated as
\begin{equation}
\label{fair-regressor}
\min_{g\in \Gc^{\text{fair}}} \Eb\Big[ \big(L - g(X,Z)\big)^2\Big];
\end{equation}
evidently, without the restriction $g\in \Gc^{\text{fair}}$, the conditional expectation $\mu(X,Z)$ would be best.
Denote by $Q(x)$ the conditional distribution of $\mu(X,Z)$, given $X=x$. The fundamental result in this context, due to
\cite{chzhen2020fair,gouic2020projection}, is that (under mild technical conditions) the optimal value of problem \eqref{fair-regressor} is equal to the squared optimal value of the barycenter problem for $p=2$:
\begin{equation}
\label{fair-barycenter}
\min_{q} \sum_{x\in \Xc}\lambda(x)\, W_2^2(Q(x),q).
\end{equation}
Furthermore, the barycenter $\bar{q}$ is equal to the distribution of $\hat{g}(X,Z)$, where $\hat{g}$ solves problem \eqref{fair-regressor}. The latter, by the definition of fairness,
is the same as the conditional distribution of $\hat{g}(X,Z)$, given $X=x$, for any $x$. This leads to the following procedure for solving problem \eqref{fair-regressor}, as
detailed in \cite{gouic2020projection}.
First, we find Bayes regressors $\mu(X,Z)$ and develop their distributions $Q(x)$, $x\in \Xc$. The second step is to find their barycenter $\bar{q}$ from
\eqref{fair-barycenter}. We can also find for each $x$ a mapping $T_x$ such that $T_x\mu(x,Z)$ has distribution $\bar{q}$, and a mapping $T^x$ such that for any variable $Y$ with distribution $\bar{q}$,
$T^x Y$ has distribution $Q(x)$. Then, for each $x$, the best fair predictor is
\[
\hat{g}(x,Z) = \sum_{x'\in \Xc} \lambda(x') T^{x'}T_x \,\mu(x,Z).
\]
Although the original motivation for this analysis was insurance, one can easily extend these techniques to other applications, such as credit qualification, housing,
health policy, education, etc. Corresponding results for classification can be found in \cite{gaucher2023fair}.
The downside of this appraoch is that the resulting mappings may be randomized.

Based on these ideas, the thesis \cite{carassai2024neural} proposes a general and robust procedure for preprocessing arbitrary data sets to guarantee fairness.
Its advantage is that it is not restricted to the least-squares method and can be applied before training by any machine learning method.
Suppose the data contains attributes and labels (combined) $(X,Z,Y)$, with $X$ being the sensitive attribute, $Z$ representing other attributes, and $Y$ being the labels. It is not sufficient to
ignore the attribute $X$ in the data before training; the result may still become unfair, due to intrinsic correlation. To make sure that any machine learning procedure will produce fair results, we replace the distribution of the data $(Z,Y)$ with the Wasserstein barycenter of the conditional distributions $Q(x)$ of $(Z,Y)$, given $X=x$, for all values of $x$.

An important aspect of health and financial applications is that the specific data distributions $Q(x)$, for $x\in \Xc$, are stored in local databases
and may not be available directly for centralized policy analysis, because of privacy or confidentiality restrictions.
We cannot just set $q$ to be the mixture distribution $\lambda \circ Q$ defined as
\[
[\lambda \circ Q](B) = \int_\Xc Q(x)(B)\;\lambda(\D x), \quad \text{for all Borel sets } B\subset \Yc,
\]
because its calculation requires direct access to the local data.

}

\subsubsection*{Existing Methods}

There are continuous efforts to develop efficient algorithms for computing relevant transportation distances that are relevant to the barycenter problem. Most work focuses on measures supported on finitely many points. The Sinkhorn algorithm, introduced by \cite{sinkhorn1}, incorporates an entropic regularization term into the mass transportation problem and has emerged as a baseline approach alongside its variant, Greenkhorn \cite{Greenkhorn}.
As observed in \cite{benamou2015iterative}, Sinkhorn is an special case of an alternating Bregman projection method. These algorithms have driven significant progress in the field \cite{sinkhorn2,APDAMD1}. Other notable approaches include accelerated primal-dual gradient descent \cite{APDAGD1,APDAGD2,APDAGD3} and semi-dual gradient descent \cite{semi_dual1,semi_dual2}.

The Wasserstein barycenter inherits the computational difficulties associated with optimal
transport distances.

{

 Problem \eqref{barycenter} exhibits features similar to two-stage stochastic programming problems (see \cite{shapiro2021lectures} and the references therein):
the barycenter distribution $q$ is the first stage variable, the distributions $Q(x)$, $x\in \Xc$, are the second-stage data, and the couplings $\pi(x)$ in \eqref{Wass} between $q$ and $Q(x)$ are the second stage variables.

Two fundamental approaches to problems of this structure are the primal and dual decomposition methods. They were developed and successfully applied
in the area of stochastic programming; see \cite[Ch. 3]{ruszczynski2003stochastic} for a thorough exposition and a historical account. Recently, similar ideas were explored
in connection with the Wasserstein barycenter problem.

In the primal approaches, the coordinating problem iterates over the first-stage variables, usually by solving a simplified model or conducting a subgradient-based search. The second stage problems are solved for each new value of the first-stage variables to provide subgradient information about their optimal values. In stochastic programming, the key contributions are
\cite{van1969shaped,ruszczynski1986regularized,birge1988multicut}. For the Wasserstein barycenter problem, a fundamental difference occurs between formulations in which the support of the barycenter is fixed, and formulations in which it is free and has to be optimized. Fixed-support models are essentially two-stage linear programming problems with transportation subproblems. For a free-support problem
\cite{cuturi2014fast} proposes a method of this class with the outer algorithm iterating over the first-stage variables (the central distribution) by a simple subgradient method, while the subproblems optimize the regularized Wasserstein distance (Sinkhorn distance) and calculate corresponding subgradients. Inaccuracies due to the regularization can be mitigated by a method of \cite{xie2020fast}.

In the dual approaches, the first-stage variables are replicated, and the subproblems optimize modified transportation problems with respect to both groups of variables. In stochastic programming, such approaches were initiated in \cite{rockafellar1991scenarios,mulvey1995new}. The purpose of coordination is to bring the copies of the
first stage variables to equality.  For the fixed-support barycenter problem, Refs. \cite{benamou2015iterative,lin2020computational} propose an approach in the spirit of the Progressive Hedging method of
\cite{rockafellar1991scenarios}: an alternating Bregman projection method, which is a direct extension of the Sinkhorn algorithm.
 Ref. \cite{dvurechenskii2018decentralize} applies a dual method, by reproducing the first stage variables, as in \cite{mulvey1995new}, and iterating over the
 dual variables by a subgradient method until coordination is achieved. Again, each iteration requires the solution of a transportation problem for each subproblem $s$.
Ref. \cite{alvarez2016fixed} proposes a fixed-point heuristic iteration.

The common drawback of the approaches to the variable-support problem is that they require an exact solution of a transportation problem to calculate the Wasserstein distance (or its modification) for each subproblem $x$ and at each iteration of the upper-level algorithm.

}

\subsubsection*{Privacy Issues and Federated Learning}

The second issue that we plan to address is the \emph{privacy of the data}.
Generally, a distributed machine learning framework that enables diverse clients or users to collaboratively train a model on a central server while preserving the decentralized nature of sensitive client data is developed in the literature under the name of
Federated Learning (FL) (see \cite{FL1,FL2,Fl3,FL_L1,FL_L2} and the references therein). A key characteristic of FL is its emphasis on \textit{statistical heterogeneity}, which refers to the non-iid   data across clients. In typical FL application, each client corresponds to a user, and the richness of their contributed data reflects various aspects, such as personal, cultural, regional, and geographical traits.

Federated learning faces several challenges that arise from its decentralized nature and the need to ensure privacy, security, and fairness. The main difficulties include \textit{communication efficiency}, as iterative communication between the center and clients can be costly; \textit{statistical heterogeneity}, due to the unique data distributions across clients; \textit{privacy and security risks}, such as inferring sensitive information from shared model updates; \textit{fairness issues}, stemming from decentralized and potentially non-iid data; and ensuring interpretability and explainability in the federated setting.  Despite the growing interest in federated learning, the above critical challenges are not fully addressed, leaving numerous ongoing research directions.

In this paper, we introduce decomposable and parallelizable dual subgradient methods for approximating client distributions and determining the Wasserstein barycenter, specifically designed for privacy-preserving settings. We refer to the algorithms based on the subgradient methods as the Federated Dual Subgradient Algorithms.
{ Our dual method is highly non-standard and it does not follow the dual decomposition approaches known from stochastic programming, such as \cite{mulvey1995new} and related research. The novel ideas are a new reformulation of the free-support problem, detailed in section \ref{s:particles}, and partial dualization of the local (subproblem) constraints, described in section \ref{s:method}. Most of the dual variables are local, and we avoid repeated solving of the mass transportation problems. Instead of that, we employ a closed-form minimization of the Lagrangian.

To the best of our knowledge, it is the first efficient methodology for determining the Wasserstein barycenter without repeated subproblem solution, and under the privacy-preserving setting.}

\section{Discrete Approximation of the Variable-Support Problem}
\label{s:particles}

In this section, the objective is to develop a convenient mixed-integer reformulation of the transportation barycenter problem, under the assumption that the spaces $\Xc$ and $\Yc$ are \emph{finite}.

Evidently, if $\Yc$ is finite, then \eqref{Wass} has a linear programming representation.
Let $\mu$ and $\nu$ be discrete measures in $\Pc(\mathcal{\Yc})$, supported at positions $\{y^{(i)}\}_{i=1}^{N}$ and $\{z^{(j)}\}_{j=1}^{S}$ with normalized (totaling 1) positive weight vectors $w_{y}$ and $w_{z}$:
\[
\mu=\sum_{i=1}^{N} w_{y}^{(i)} \delta_{y^{(i)}}, \quad \nu=\sum_{j=1}^{S} w_{z}^{(j)} \delta_{z^{(j)}}.
\]
For $p \geq 1$, let $D \in {R}_{+}^{N \times S}$ be the distance matrix with the entries $D_{i j}=d\big(y^{(i)},z^{(j)}\big)^{p}$. Then the $p$th power of the $p$-Wasserstein distance between the measures $\mu$ and $\nu$ is the optimal value of the following transportation problem:
\begin{equation} \label{12}
\min _{\pi \in {R}_{+}^{N \times S}}\  \sum_{i=1}^N\sum_{j=1}^S  D_{i j} \pi_{i j} \quad
\text {s.t.} \quad  \pi^\top \1_{N}=w_{y}, \quad  \pi \1_{S}=w_{z}.
\end{equation}
We plan to exploit this linear programming formulation within our approach.

The finiteness of $\Xc$ may result from considering a
discrete distribution $\widetilde{\lambda}$: an approximation of the true marginal distribution $\lambda$. In our approach, we represent the support of $\widetilde{\lambda}$ with a set of state points $\big\{x^s\big\}_{s=1,\dots,N}$; this may be, for example, a large sample from $\lambda$. Furthermore, each distribution $Q(x^s)$ is represented by a finite number of particles $\big\{y^{s,i}\big\}_{i\in \Ic^s}$ drawn independently from $Q(x^s)$. As every conditional distribution $Q(x^s)$
is private, we assume that the sample  $\big\{y^{s,i}\big\}_{i\in \Ic^s}$ is only available locally, to the ``device'' $s$, but not to other devices or to the coordinator. We use the symbol $\widehat{Q}(x^s)$
to denote the empirical distribution supported on $\big\{y^{s,i}\big\}_{i\in \Ic^s}$, and thus $\widehat{Q}$ is a sample-based kernel closely approximating $Q$, but subject to the same privacy restrictions. Refs. \cite {dereich2013constructive} and \cite{fournier2015rate} provide tight estimates
on the accuracy of empirical approximations of probability measures, { and Ref. \cite{heinemann2022randomized} guarantees for the resulting approximation of the barycenter problem.}

%

To achieve the finiteness of the target space, we consider
a set $\Zc =\big\{\zeta^k\big\}_{k=1,\dots,K} \subset \Yc$.
It consists of pre-selected potential locations for the support of the measure $\bar{q}$ sought in \eqref{barycenter}. For example, the set
$\Zc$ may be a large sample from the mixture distribution $\lambda\circ \widetilde{Q}$, where $\widetilde{Q}$ is a model or approximation of the unknown $Q$, or a dense grid in $\Yc$. The cardinalities of the finite sets introduced here: $N$ and
$K$, may be very large; our methods will depend on them in a linear way only.

At this stage, we depart from the usual formulations of a free-support barycenter problem.
We restrict our search to uniform discrete measures $\bar{q}$ which have their support of cardinality $M \ll K$ within the set $\Zc$. Thus, we aim to find an \textit{ensemble of particles} $\widetilde{\Zc} \subset \Zc$, such that
$|\widetilde{\Zc}|= M$ and the measure
\begin{equation}
    \label{qZ}
\bar{q}(\widetilde{\Zc}) = \frac{1}{|\widetilde{\Zc}|} \sum_{y\in \widetilde{\Zc}}\delta_y
\end{equation}
is the best in \eqref{barycenter}
among all measures of this structure.  Formally, we define the following discrete approximation of problem \eqref{barycenter}:
\begin{equation}
\label{barycenter-seearch}
\min_{{\widetilde{\Zc}\subset \Zc}\atop{|\widetilde{\Zc}| = M}}
\left\{\Wc_{p}^{\widetilde{\lambda}}(\widehat{Q}, \bar{q}(\widetilde{\Zc}))^p=
\sum_{s=1}^N \widetilde{\lambda}_s
\big[{W}_{p}\big(\widehat{Q}(\,\cdot\, | x^s), \bar{q}(\widetilde{\Zc})\big)\big]^p
\right\}.
\end{equation}
{ Our problem is a discretized version of the notoriously difficult variable-support barycenter problem.}

Problem \eqref{barycenter-seearch} is a combinatorial optimization problem of considerable complexity, and a brute-force approach to it is doomed.
We will convert it into a mixed-integer linear programming problem, which still appears extremely difficult but has a structure that can be exploited to develop a rapid solution method preserving privacy restrictions.

First, we introduce the binary variables
\[
\gamma_k = \begin{cases} 1 & \text{ if the point $\zeta^k$ has been selected to $\widetilde{\Zc}$},\\
0 & \text{ otherwise},
\end{cases}
\quad k=1,\dots,K.
\]
With this notation, for a nonzero $\gamma$, the measure \eqref{qZ} has the form:
\begin{equation}
\label{qZ-b}
\bar{q}(\widetilde{\Zc})_k = \frac{\gamma_k}{\sum_{j=1}^K \gamma_j}, \quad k=1,\dots,K.
\end{equation}
We may remark here that the above formula defines a probability measure $\bar{q}$ for any nonzero and nonnegative vector $\gamma$.

Next, we delve into the linear programming formulation \eqref{12}
of the Wasserstein distance: for every $s$ and every nonzero
binary vector $\gamma$, the quantity $\big[{W}_{p}\big(\widehat{Q}( x^s), \bar{q}(\widetilde{\Zc})\big)\big]^p$ is the optimal value
of the following transportation problem:
\begin{equation}
\label{pi-gamma-problem}
\begin{aligned}
\min \ & \sum_{i\in \Ic^s}\sum_{k=1}^K d_{sik}\pi_{sik}\\
\text{s.t.} \
&\sum_{i\in \Ic^s} \pi_{sik} = \frac{\gamma_k}{\sum_{j=1}^K \gamma_j}, \quad k=1,\dots,K,\\
&\sum_{k=1}^K \pi_{sik} = \frac{1}{|\Ic^s|}, \quad i\in \Ic^s,\\
&\pi_{sik} \ge 0, \quad i\in \Ic^s, \quad k=1,\dots,K,
\end{aligned}
\end{equation}
with $d_{sik}=d\big( y^{s,i} , \zeta^k\big)^p$.
Observe that in this formulation we do not assume that $\sum_{j=1}^K \gamma_j=M$, although this equation is required at the solution. In our method,
 we shall deal with intermediate approximate solutions violating this cardinality constraint, and we shall need a precise formulation of the
 Wasserstein distance in these cases as well.

 The first constraint of problem \eqref{pi-gamma-problem}, requiring that
 the $\Yc$-marginal of the transportation plan $\varPi$ matches the  distribution \eqref{qZ-b}, is linear with respect to
 $\pi$, but not with respect to the variables  $\gamma$. To remedy this
 problem, we re-scale the transportation plans by introducing new variables:
\begin{equation}
    \label{pi-beta}
\beta_{sik} = \pi_{sik} {\sum_{j=1}^K \gamma_j}, \quad s=1,\dots,N,\quad  i \in \Ic^s,\quad  k=1,\dots,K.
\end{equation}

The following linear mixed-integer optimization problem
integrates
problems \eqref{pi-gamma-problem} into~\eqref{barycenter-seearch}:
\begin{subequations}
\label{mixed-bin-b}
\begin{align}
\min_{\gamma,\beta}
&\; { \sum_{s=1}^N  w_s \sum_{i\in \Ic^s}\sum_{k=1}^k d_{sik} \beta_{sik} }\label{mb-a}\\
\text{s.t.}
&\; \beta_{sik}\ge 0,\  \gamma_k\in \{0,1\}, \quad s=1,\dots,N,\quad  i \in \Ic^s,\quad  k=1,\dots,K,\label{mb-e}\\
&\;\sum_{i\in \Ic^s} \beta_{sik} = \gamma_k,\quad s=1,\dots,N,\quad k=1,\dots,K, \label{mb-b}\\
&\; {\sum_{k=1}^K \beta_{sik} =  \frac{1}{|\Ic^s|}\sum_{k=1}^K\gamma_k , \quad s=1,\dots,N,\quad i \in \Ic^s}, \label{mb-c}\\
&\; {\sum_{k=1}^K \gamma_k  = M},\label{mb-d}
\end{align}
\end{subequations}
with $w_s = \frac{\widetilde{\lambda}^s}{M}$. In \eqref{mb-a}, we undo the change of variables \eqref{pi-beta}, using $M$ as the normalization constant. with a view to \eqref{mb-d}. Still, the constraints \eqref{mb-b}--\eqref{mb-c} correctly model the mass transportation for any nonzero $\gamma$. { We do not substitute
$M$ for $\sum_{k=1}^K \gamma_k$ in \eqref{mb-c} on purpose, to facilitate our dual method to be described in the next section.

Again, if the binary restrictions on $\gamma$ are relaxed
to the requirement that $\gamma_k\in [0,1]$, $k=1,\dots,K$,
the problem \eqref{mixed-bin-b} correctly models the
fixed-support barycenter problem, with the restriction of the
barycenter support to $\Zc$. Its optimal value is the $p$th power of the optimal value of \ref{barycenter}, with this additional restriction.}

\section{A Federated Dual Subgradient Method}
\label{s:method}

As problem \eqref{mixed-bin-b} involves binary variables, it might be tempting to employ an integer programming solver, such as Gurobi, CPLEX, or SCIP. However, two issues or limitations exist for such integer programming solvers. First, integer or even linear programming can become computationally intractable for large-scale problems with numerous support points of the measures. Additionally, to formulate such problems with mixed-integer programming, we would need to transmit data from all local devices to the global device  (coordinator) and use it to train a global model that would incorporate information from all local devices' data. Therefore, to address the limitations of computational intractability, privacy,  and security,   we propose dedicated subgradient-based decomposition methods to solve problem \eqref{mixed-bin-b}.

{  Problem \eqref{mixed-bin-b} still has the structure of a two-stage problem of stochastic programming: the variables $\gamma_k$, $k=1,\dots,K$, are the first-stage variables, which define the
barycenter distribution \eqref{qZ-b}, while the variables $\beta_{sik}$, $s=1,\dots,N$, $i\in \Ic_s$, are the second-stage variables, defining the couplings $\pi_s$ between the barycenter and the empirical distributions for each subproblem~$s$.
However, we do not follow the ideas of the decomposition methods discussed above, but rather develop a completely new dual method exploiting the
specificity of the Wasserstein barycenter problem. The main idea is to avoid solving the exact mass transportation subproblems \eqref{pi-gamma-problem}
for each value of the upper level variables $\gamma$ (determining the current approximation of the barycenter distribution). Instead of that, we solve simple Lagrangian relaxations
based on multipliers associated with the assignment equations \eqref{mb-c}. Owing to that, the approximate transportation problems have closed-form solutions.
The multipliers are iterated in parallel with the upper-level variables (a single-loop approach) and converge to the correct values, so that exact transportation distances are
calculated at termination.
}

We work with central primal variables $\gamma_k$, $k=1,\dots,K$, and one dual variable updated by the coordinator, and groups of local primal and dual variables which are
updated by the clients. In this way, we obtain the desired privacy of the local data.

Assigning Lagrange multipliers $\theta_{si}$ and $\theta_0$ to the constraints \eqref{mb-c} and \eqref{mb-d}, respectively,
we obtain the Lagrangian function of problem \eqref{mixed-bin-b}:
\[
 L(\gamma,\beta;\theta)
 =  \sum_{s}\sum_{i}\sum_{k} w_s d_{sik} \beta_{sik}
 + { \sum_{s}\sum_{i}\theta_{si} \sum_{k} \big(\frac{\gamma_k}{|\Ic^s|} -  \beta_{sik} \big)}   +\theta_{0} \big(\sum_{k} \gamma_k -  M  \big).
\]
The dual variable $\theta_0$ has the interpretation of the marginal contribution of an additional point of the cloud to reducing the distance \eqref{mb-a}. The variables
$\theta_{si}$ serve as thresholds in the assignment of the particles $y^{s,i}$ to the candidate points. We designate $\theta_0$ as the \textit{global} dual variable, accessible to both the central device and the local devices. Conversely, $\theta_{si}$, $i\in \Ic^s$, are \textit{local} variables: they are exclusively accessible to the local device \(s\). The central device will only need a $k$-dimensional vector from each device $s$.

The corresponding dual function is
\begin{align}
\lefteqn{L_D(\theta)  = \min_{\gamma,\beta \in \Gamma} L(\gamma,\beta; \theta)} \label{dual_barycenter}\\
 &= \sum_{k=1}^{K} \, \min_{\gamma_k,\beta_{\cdot\cdot k} \in \Gamma_k} \, \bigg\{  \sum_{s=1}^{N}\sum_{i\in \Ic^s}(w_s d_{sik} - \theta_{si})\beta_{sik} + \Big(\sum_{s=1}^{N}\frac{1}{|\Ic^s|}\sum_{i\in \Ic^s}\theta_{si} + \theta_{0}\Big)\gamma_k \, \bigg\}   - M\theta_{0},
 \notag
\end{align}
where $\Gamma$ is the feasible set of the primal variables given by the conditions \eqref{mb-e}--\eqref{mb-b}, and $\Gamma_k$ is its projection on the
subspace associated with the $k$th candidate point $\zeta^k$.


The minimization in \eqref{dual_barycenter} decomposes into $K$ subproblems, each having a closed-form solution.
{ First, we determine how the local assignment variables $\beta_{sik}$ depend on the $\gamma_k$'s.}
 From the definition of the set $\Gamma_k$, we see that if $\gamma_k=0$ then  $\beta_{sik} = 0$ for all $s=1,\dots,N$ and $i\in \Ic^s$. If $\gamma_k=1$ then for every $s$ it is best to make
$\beta_{sik}=1$ for $i = i^*(s,k)$ for which the difference $w_s d_{sik} - \theta_{si}$ is the smallest. { Other $\beta_{sik}=0$, for $i\ne i^*(s,k)$.}
Then
\[
\sum_{i\in \Ic^s}(w_s d_{sik} - \theta_{si})\beta_{sik} = \min_{i\in \Ic^s}\;\{w_s d_{sik} - \theta_{si}\}.
\]
{ This allows us to determine the best value of each $\gamma_k$ by simply looking at the two possible values of the expression in braces in the formula
\eqref{dual_barycenter}.}
For all $k=1,\dots,K$,
\begin{equation}
    \label{gamma-opt}
\gamma_k  =  1, \quad  \text{ if }\quad  \sum_{s=1}^{N}\frac{1}{|\Ic^s|}\sum_{i\in \Ic^s}\theta_{si} + \theta_{0} < \displaystyle{\sum_{s=1}^{N}\max_{i\in \Ic^s}}\;\{ \theta_{si} -w_s d_{sik}\};
\end{equation}
$\gamma_k \in\{0,  1\}$, if exact equality holds; and $\gamma_k=0$, otherwise.
{ This determination can be made on the basis of the values of the quantities
\begin{equation}
\label{reported}
T_{sk} = \max_{i\in \Ic^s}\{ \theta_{si} -w_s d_{sik}\} - \frac{1}{|\Ic^s|}\sum_{i\in \Ic^s}\theta_{si},\quad k=1,\dots,K,
\end{equation}
received from the subproblems $s=1,\dots,N$. The couplings $\beta$ then follow as described above.}

The dual problem has the form
\begin{equation}
\label{dual-problem-barycenter}
\max_{\theta} L_D(\theta).
\end{equation}

The subdifferential of the dual function can be calculated in the standard way:
{
\begin{equation}
\label{subdifferential-barycenter}
\partial L_D(\theta)=\operatorname{conv}\left\{ \begin{bmatrix}
\left\{\sum_{k=1}^K \big(\frac{\hat{\gamma}_k}{|\Ic^s|} -  \hat{\beta}_{sik} \big) \right\}_{s=1,\dots,N,\  i \in \Ic^s }\\
\sum_{k=1}^{K} \hat{\gamma}_k - M \end{bmatrix}: (\hat{\gamma},\hat{\beta})\in \hat{\Gamma}(\theta) \right\},
\end{equation}
where $\hat{\Gamma}(\theta)$ is the set of solutions of
problem \eqref{dual_barycenter}.
}
At the optimal solution $\hat{\theta}$ we have $0 \in \partial L_D(\hat{\theta})$.

As usual with the dual methods for mixed-integer programming, the optimal value of \eqref{dual-problem-barycenter} may be strictly below the optimal value of \eqref{mixed-bin-b};
it is equal to the optimal value of the linear programming relaxation, where the conditions $\gamma_k\in\{0,1\}$ are replaced by $\gamma_k\in [0,1]$. { However, such relaxation
of \eqref{mixed-bin-b} correctly models the barycenter problem, with the support of $q$ restricted to $\Zc$.}

{ An essential virtue of our dual method is that the rule \eqref{gamma-opt} naturally defines the $\gamma_k$'s as binary. The number of selected points may differ from $M$, but $M$ is not a hard parameter of the model.}
If we replace $M$ by the number of $\gamma_k$'s equal to 1, the gap is zero; the solution of the relaxation is optimal for the mixed integer formulation. If we keep $M$ unchanged, we can construct a feasible solution by setting to 0 the $\gamma_k$'s for which the change in the expression in the braces in \eqref{dual_barycenter} is the smallest. This allows for the estimation of the gap. { But this correction is not really necessary.}
It is worth stressing that in the algorithm below, we need only \emph{one} subgradient for each~$\theta$, and thus one minimizer in \eqref{dual_barycenter}.


\section{Single-Loop Algorithms for Federated Learning}
\label{s3.2}

In the FL setting, the $\gamma$'s are global variables (known to both the central device and the local devices). The dual variables $\theta_{si}$ and the $\beta_{si}$'s are private or local variables (known only to the corresponding devices $s$). The central device assumes the responsibility for iterating the dual variable $\theta_0$ by aggregating the numbers reported by the local devices.

{ An iteration of the method does not involve solving any
mass transportation problems, but only closed-form operations.}
Each local device calculates its temporary values of the $\beta$'s (by setting $\beta_{s,i^*(s,k),k}=1$,  and other to 0) and
reports the quantities \eqref{reported}
to the global device. Upon receiving this information, the global device aggregates the quantities \eqref{reported}, determines the value of $\gamma$ by \eqref{gamma-opt}, makes a subgradient step in $\theta_0$, and relays it back to the local devices. Subsequently, each local device updates its $\theta_{si}$'s using the $\gamma$'s and (corrected) $\beta$'s in a subgradient step. Then the iteration continues.

By employing the dual subgradient method, the central device avoids using private data, ensuring privacy preservation. It does not know the data points $y^{s,i}$, it does not know the distances $d_{sik} = d(y^{si},\zeta^k)^p$, it does not know the local dual variables $\theta_{si}$, and not even the local numbers of particles $|\Ic^s|$ or
the weights $w_s$. The only local information reported is \eqref{reported}.

The calculations do not involve any matrix-vector operations. For each device $s$, and at each iteration of the method, only one pass over source-destination pairs must be executed. The work of the global device is of the
order $\mathcal{O}(N\cdot K)$, where $N$ is the number of devices, and
$K$ is the total number of potential reference points.

The communication cost per iteration, representing the amount of data that needs to be transmitted between the central server and the participating clients (devices or edge nodes) for each local device \(s = 1,\ldots, N\) is of the order of  \(\mathcal{O}( K)\).

In Algorithm \ref{a-barycenter} below, the variable \(j\) denotes the iteration number, starting from 0. The symbol \(\theta\) represents the initial values of the dual variables, while \(M\) is the desired cardinality of the support of the barycenter. The parameter \(\epsilon\) defines the tolerance level, and \(\alpha^{(0)}\) represents the initial learning rate. The variables \(\varkappa_1\) and \(\varkappa_2\) are exponential decay factors between 0 and 1, determining the relative contribution of the current subgradient and earlier subgradients to the direction. The variable ``maxiter'' refers to the maximum number of iterations allowed. It's important to note that the total number of \(\gamma\)'s selected by the subgradient method may not necessarily equal \(M\) when the stopping criteria are met. We introduce the parameter \(a\) to ensure that the iteration concludes when \(\sum_{k=1}^{K} \gamma_k\) is close to \(M\). Additionally, the global steps and local steps, performed by the global and local devices, respectively, are indicated by comment lines starting with ``\textbackslash\textbackslash''. The global variables $\gamma$ are accessible to the local devices.


\begin{algorithm}[h!]
\caption{ Federated dual subgradient algorithm for the barycenter problem \eqref{mixed-bin-b}}\label{a-barycenter}
\begin{algorithmic}[1]
\Require{$\theta^{(0)}$, $M$, $\epsilon$, $\alpha^{(0)}$,$\varkappa_1$,$\varkappa_2$,maxiter and $j=0$.}
\Ensure{ $\theta$,$\gamma$, and $\beta$.}
\While{$j < $ maxiter}
 \State \textbackslash \textbackslash \textbf{\emph{The local devices $s = 1, \dots, N$}}
 \State $T^1_{s} \leftarrow \frac{1}{|\Ic^s|}\sum_{i\in \Ic^s}\theta_{si}^{(j)}$
 \For{$k = 1, \dots, K$}
 \State $T^2_{sk} \leftarrow \max_{i\in \Ic^s} (\theta_{si}^{(j)} -w_s d_{sik})$,  and $T_{sk} \leftarrow T^2_{sk} - T^1_{s}$
 \EndFor
 \State Report  $T_{sk}$, $k=1,\dots,K$, to the global device

 \State \textbackslash \textbackslash \textbf{\emph{The global device}}
 \For{$k = 1, \dots, K$}
 \If{$\sum_{s=1}^{N}T_{sk} > \theta_0^{(j)}$  }
\State $\gamma_k \leftarrow 1$\;
 \Else
   \State $\gamma_k \leftarrow 0$\;

 \EndIf
 \EndFor

 \State The global device computes $L(\theta^{(j)})$ using eq. \eqref{dual_barycenter}

   \If{$\| L_D{(\theta^{(j)})} - L_D{(\theta^{(j-1)})} \| \leq \epsilon $ }
\State break from the while loop

   \EndIf

  \State$ \alpha^{(j+1)} \leftarrow \frac{\alpha^{(0)}}{\sqrt{j+1}}$
   \State $m^{(j+1)}_0 \leftarrow (1-\varkappa_1)(\sum_{k=1}^{K} \gamma_k-M) + \varkappa_1m^{(j)}_{0}$\;
  \State  $\theta^{(j+1)}_{0} \leftarrow \theta^{(j)}_{0} + \alpha^{(j+1)} m^{(j+1)}_{0}$\;
   \State Send $\gamma_{k}$, $k=1,\dots,K$, to the local devices

  \State \textbackslash \textbackslash \textbf{\emph{The local devices  $s=1,\ldots,N$.}}
  \For{$k = 1, \dots, K$}
  \If{$\gamma_k = 1$}
  \If{$i = \argmax_{i\in \Ic^s}\;\{\theta_{si}^{(j)} - w_s d_{sik}\}$  }
\State $\beta_{sik} \leftarrow 1$\; \textbackslash \textbackslash If multiple maximizers exist, we  randomly select one $i$.
 \Else
   \State $\beta_{sik} \leftarrow 0$\;
   \EndIf
   \EndIf
 \EndFor
 \State  $m^{(j+1)}_{si} \leftarrow (1-\varkappa_2)\sum_{k=1}^K (\frac{\gamma_k}{|\Ic^s|} -  \beta_{sik}  * \gamma_k) + \varkappa_2 m^{(j)}_{si}$\;
  \State  $\theta^{(j+1)}_{si} \leftarrow \theta^{(j)}_{si} + \alpha^{(j+1)} m^{(j+1)}_{si}$ $ \quad  s = 1, \dots, N$, $i\in \Ic^s$\;

  \State  $j \leftarrow j + 1\; $
 \EndWhile
\end{algorithmic}
\end{algorithm}

In the context of (approximate) primal recovery, we select the values \(\theta^{(j)}\) for the last \(J\) iterations where \(L_D(\theta^{(j)})\) closely approaches the optimal solution. Subsequently, we construct the convex hull of the observed subgradients of the dual function at these points, approximating the subdifferential defined in equation \eqref{subdifferential-barycenter}. The minimum norm element within this convex hull represents a convex combination of the corresponding dual points: \((\bar{\gamma},\bar{\beta}) = \sum_{j\in J} \omega_j(\gamma^{(j)},\beta^{(j)})\), with \(\sum_{j\in J} \omega_j = 1\) and \(\omega_j \ge 0\). Based on convex optimization duality theory, if the subgradients were obtained at the optimal point, \((\bar{\gamma},\bar{\beta})\) would serve as the solution to the convex relaxation of problem \eqref{mixed-bin-b}. Hence, if the norm of the convex combination of the subgradients is small, we can accept \(\bar{\gamma}\) as an approximate solution. We interpret this as the optimal "mixed strategy" and select each point \(k\) with a probability proportional to \(\bar{\gamma}_k\). In our experiments, we adopt a straightforward strategy using \(\omega_j = \big( \sum_{i\in J}\alpha^{(i)}\big)^{-1}\alpha^{(j)}\approx 1/|J|\). This approach finds theoretical support in \cite{larsson1999ergodic}. The convergence rate for the subgradient method, \(\mathcal{O}(1/\sqrt{j+1})\), has been well-established since \cite{zinkevich2003online} (also referenced in \cite{garrigos2023handbook}).

In the stochastic adaptation of the approach, the iteration over $k$ is conducted within a randomly chosen batch $\Bc^{(j)} \subset \{1,\dots,K\}$ of size $B\ll K$. Subsequently, the subgradient component $g_0^{(j)} = \sum_{k=1}^{K} \gamma_k-M$ in line 26 is replaced by its stochastic approximation $\tilde{g}_0^{(j)} = (K/B)\sum{k\in \Bc^{(j)}} \gamma_k-M$. Likewise,  the subgradient components $g_{si}^{(j)} = 1-\sum_{k=1}^{K} \beta_{sik}$
in line 30 are substituted with their estimates $\tilde{g}_{si}^{(j)} = 1-(K/B)\sum_{k\in \Bc^{(j)}} \beta_{sik}$. If the batches are randomly drawn at each iteration, independently of all other data, the algorithm can be seen as a version of the stochastic subgradient method with momentum, as outlined in \cite{yan2018unified,liu2020improved}, and related literature.

{
Despite the algorithms' privacy-preserving design, there may still be privacy risks if the implementation is not carefully executed. Inadvertent leakage of sensitive information or the possibility of reconstructing individual data from shared model updates or gradients could compromise user privacy, especially in scenarios involving small samples of highly sensitive data. Furthermore, with non-IID data distributions across clients, imbalances in client participation, data skewness, or unequal contributions from different clients could lead to biased representations in the computed barycenter, potentially undermining fairness and accuracy.
}

\section{Numerical Illustration}

 We use three simple numerical examples to demonstrate the application of the Federated Dual Subgradient Algorithm to approximate Wasserstein barycenters under privacy-preserving settings. In each example, we analyze Gaussian mixture models (GMM) composed of $N$ Gaussian distributions. Each Gaussian component is characterized by different weights, representing their respective contributions to the overall mixture. The objective is to determine the Wasserstein barycenter of the GMMs, which represents a central tendency or average distribution that minimizes the Wasserstein distance with respect to all the component distributions.  This experiment with the 2-dimensional GMMs aims to illustrate the potential of employing the dual algorithm to determine the Wasserstein barycenter in a privacy-preserving setting.

\begin{figure}[!htbp]
\centering
\begin{subfigure}{0.45\textwidth}
{\includegraphics[width=\textwidth]{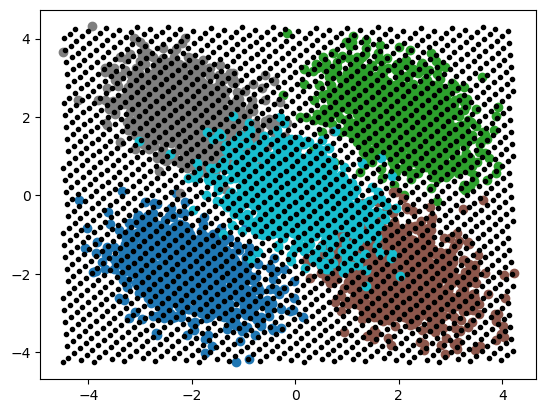}}
\end{subfigure}
\begin{subfigure}{0.45\textwidth}
{\includegraphics[width=\textwidth]{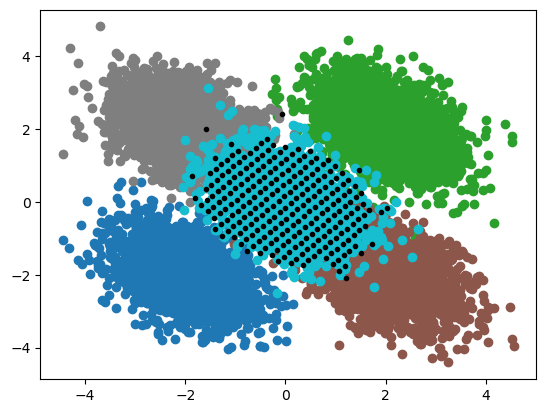}}
\end{subfigure}
\begin{subfigure}{0.45\textwidth}
{\includegraphics[width=\textwidth]{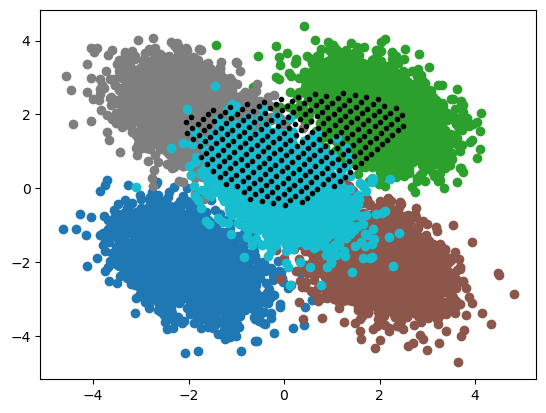}}
\end{subfigure}
\begin{subfigure}{0.45\textwidth}
{\includegraphics[width=\textwidth]{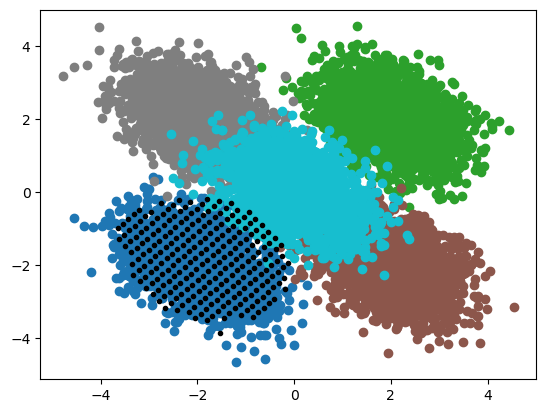}}
\end{subfigure}
\caption{The GMMs with 5 components and the selected barycenter distribution.}
\label{fig:GMM}
\end{figure}

\begin{figure}[!htbp]
\centering
\begin{subfigure}{0.45\textwidth}
{\includegraphics[width=\textwidth]{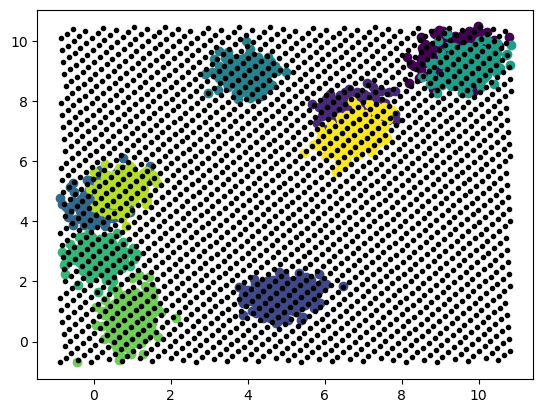}}
\end{subfigure}
\begin{subfigure}{0.45\textwidth}
{\includegraphics[width=\textwidth]{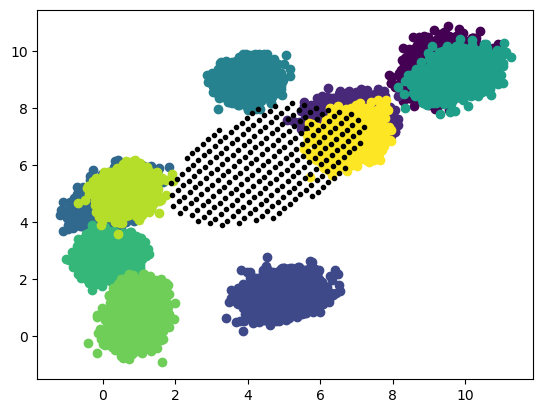}}
\end{subfigure}
\caption{The GMMs with 10 components and the selected barycenter distribution.}
\label{fig:GMM10}
\end{figure}

In the following experiment, two-dimensional Gaussian mixture models (GMMs) are composed of five Gaussian distributions. The means and covariance matrices are provided below. Figure \ref{fig:GMM} illustrates the samples of 500 particles representing the GMMs,  each in a different color, alongside the 2000-point grid of the potential barycenter support points  (marked in black) in the top-left corner. The subsequent three figures depict the GMMs compared to the selected barycenters with support cardinality $M=500$. The weights assigned to the GMMs in these figures vary: $w = [0.2,0.2,0.2,0.2,0.2]$ (top-right); $w = [0.166, 0.385, 0.063, 0.321, 0.065]$ (bottom-left); and $w= [0.7,0.1,0.05,0.05,0.1]$ (bottom-right). As observed, with changes in the weights, the approximated barycenter shifts towards the more heavily weighted centers. Additionally, Figure \ref{fig:GMM10} depicts the selected Wasserstein barycenter for a 2-dimensional GMM with 10 Gaussian components. The weight, mean, and covariance matrix of each Gaussian component are all randomly generated; the number of particles was the same as before.
 \begin{gather*}
\mu_1 = \begin{bmatrix} -2 \\ -2 \end{bmatrix}, \quad
\mu_2 = \begin{bmatrix} 2 \\ 2 \end{bmatrix}, \quad
\mu_3 = \begin{bmatrix} 2 \\ -2 \end{bmatrix}, \quad
\mu_4 = \begin{bmatrix} -2 \\ 2 \end{bmatrix}, \quad
\mu_5 = \begin{bmatrix} 0 \\ 0 \end{bmatrix}.\\
\Sigma_i = \begin{bmatrix} 0.5 & -0.2 \\ -0.2 & 0.5 \end{bmatrix}, \quad i = 1,\ldots,5.
\end{gather*}

However, we quickly recognize that issues such as data leakage or privacy concerns may arise with specific choices of hyperparameters and a fairly accurate reference set. As Figures \ref{fig:GMM} and \ref{fig:GMM10} demonstrate, the global device may capture the essence of the data distributions of local devices with minimal errors, if the weights are skewed.

\begin{figure}[!htbp]
\centering
\begin{subfigure}{0.45\textwidth}
{\includegraphics[width=\textwidth]{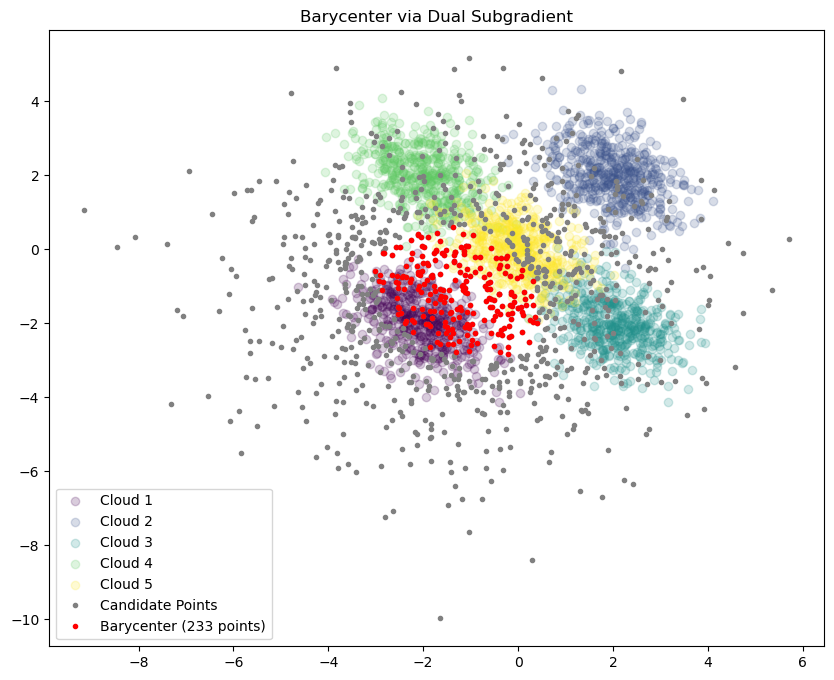}}
\end{subfigure}
\begin{subfigure}{0.45\textwidth}
{\includegraphics[width=\textwidth]{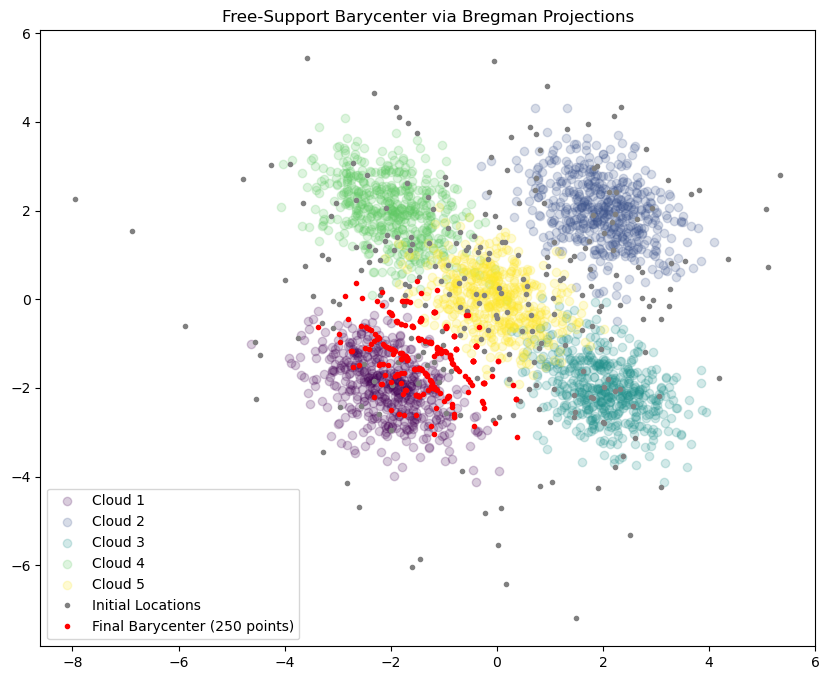}}
\end{subfigure}
\begin{subfigure}{0.45\textwidth}
{\includegraphics[width=\textwidth]{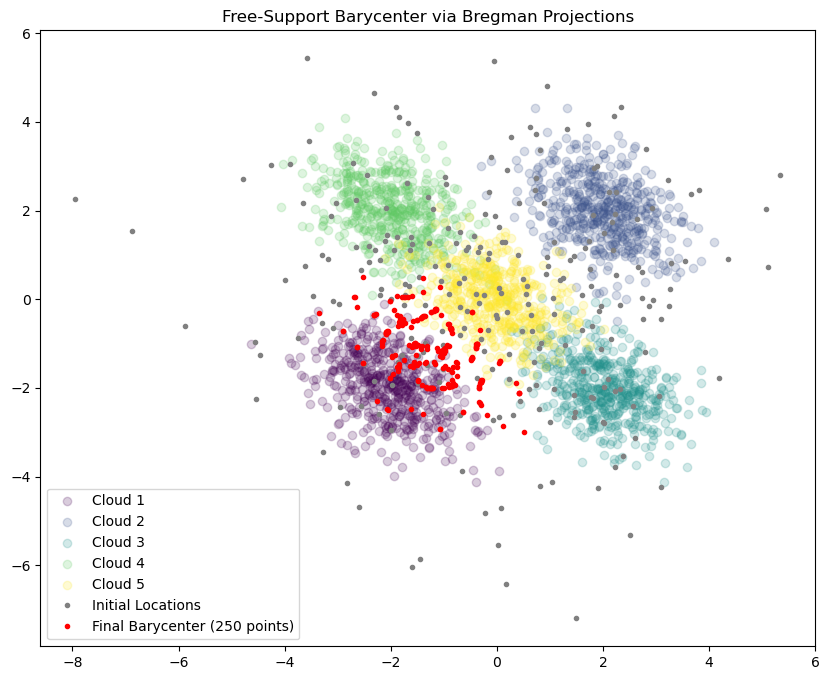}}
\end{subfigure}
\begin{subfigure}{0.45\textwidth}
{\includegraphics[width=\textwidth]{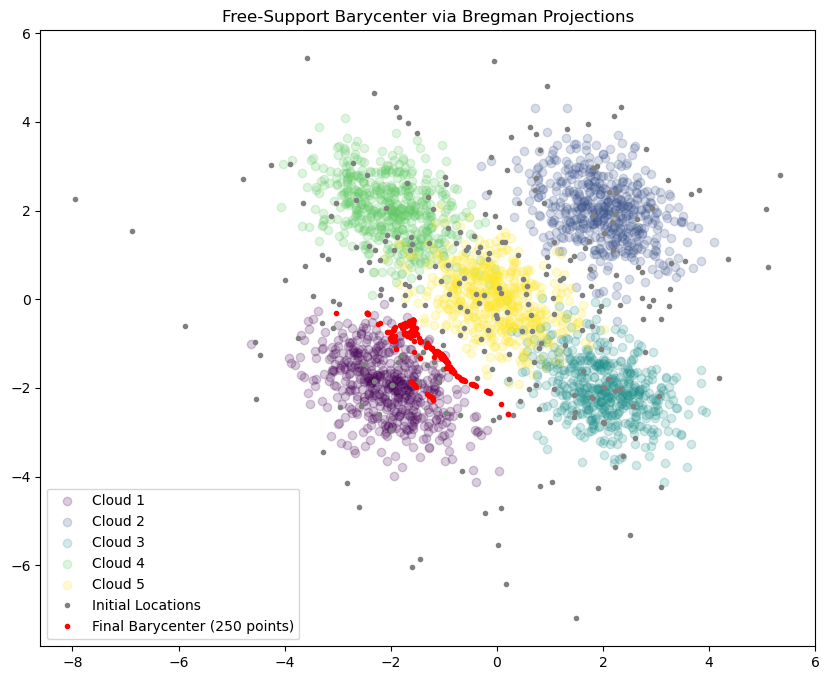}}
\end{subfigure}
\caption{Barycenters obtained by the different methods: dual subgradient method (top-left), Bregman projections with Sinkhorn regularization term 0.05 (top-right), 0.1 (bottom-left), and 0.5 (bottom-right). }
\label{fig:barycenter_methods}
\end{figure}

{
We also present an experiment to compare two algorithms for computing the Wasserstein barycenter. Our proposed dual subgradient method is a discretized free-support approach, where the barycenter's support is a subset of a larger, pre-defined set of candidate points. In contrast, the baseline method is a free-support algorithm based on Bregman projections, which optimizes the locations of a fixed number of barycenter points. This iterative process computes regularized optimal transport plans to each source measure via the Sinkhorn algorithm at each step, subsequently updating the support locations toward the barycentric mapping of the measures until convergence.
The method was adapted from the {Python Optimal Transport (POT)} library, a free collection of state-of-the-art algorithms for mass transportation problems, including the barycenter problem.\footnote{The list of contributors is provided at
{\tt https://pythonot.github.io/contributors.html}.} To ensure a rigorous comparison, both algorithms are initialized with identical source distributions, and their performance is evaluated on computational efficiency and solution quality. Efficiency is quantified by the total solution time and the average time per iteration for each method. The quality of the solution is assessed quantitatively by calculating the final Wasserstein barycenter objective value and qualitatively through a visual illustration of the structure of the resulting barycenter.

The visualization of barycenters in Figure \ref{fig:barycenter_methods} is similar to the one presented earlier in Figure \ref{fig:GMM}, but focuses on comparing the dual subgradient method with the free-support method using fixed-point iterations with alternating Bregman projections with different regularization terms.  The two-dimensional Gaussian mixture models (GMMs) consist of five Gaussian components, with the same means and covariance matrices as described above. The $K=1000$ candidate points for the dual subgradient method were randomly sampled from a normal distribution with independent components having zero means and variances equal to 5. The starting support of cardinality $M=250$ of the barycenter for the Bregman projection method was sampled from the same distribution. The stopping test for the dual method is whether the relative change in the dual function \eqref{dual_barycenter} is smaller than $\varepsilon = 10^{-4}$ and the number of points selected is within $\pm 10\%$ of $M=250$. For the alternating Bregman projection method, the stopping test is whether the relative change
of the barycenter drops below $\varepsilon = 10^{-4}$.

The subsequent figures illustrate the GMMs and their corresponding barycenters. The weights assigned to the GMMs are  $w = [0.7, 0.1, 0.05, 0.05, 0.1]$. The barycenter methods visualized are: the dual subgradient method (top-left), alternating Bregman projections with Sinkhorn regularization term 0.05 (top-right), 0.1 (bottom-left), and 0.5 (bottom-right). Table~\ref{tab:barycenter_comparison} provides the numbers of iterations, solution time, and the final barycenter value.

A key insight from this analysis is the rapid convergence of the dual method to a high-quality solution and the fundamental trade-off between accuracy and efficiency in Sinkhorn-based barycenter computations. According to Figure \ref{fig:barycenter_methods}, a smaller regularization term produces a more accurate approximation of the true Wasserstein distance, but this comes at the cost of significantly slower convergence for the Sinkhorn algorithm. When we set the regularization term to be $0.05$, the algorithm did not reach the tolerance threshold $\epsilon =10^{-4}$ before the maximum iterations.
In contrast, the dual subgradient method demonstrates superior computational performance and scalability, and much lower time per iteration. Its primary advantage lies in its architecture, which does not require the expensive computation of an optimal transport plan in each iteration. By relying on computationally cheaper dual variable updates, the method is substantially faster per iteration. This efficiency makes it more scalable as the problem size, such as the number of input distributions or the number of candidate points, increases, allowing it to outperform even the heavily regularized (e.g., 0.5) Sinkhorn-based approaches.
}

\begin{table}[h!]
\centering
\caption{Performance Comparison of Barycenter Computation Methods}
\label{tab:barycenter_comparison}
\begin{tabular}{l c c c c c c}
\toprule
\textbf{Method} & \hspace{-.3em}\textbf{Regularization}\hspace{-.3em} & \hspace{-.3em}\textbf{Converged?}\hspace{-.3em} &\hspace{-.3em}\textbf{Total Time (s)}\hspace{-.3em} & \hspace{-.3em}\textbf{Iterations}\hspace{-.3em} & \hspace{-.3em}\textbf{Time/Iter. (ms)}\hspace{-.3em} & \hspace{-.3em}\textbf{Barycenter Value}\hspace{-.3em} \\
\midrule
Dual Subgradient      & N/A    & Yes & 23.73 
& 2204 
& 10.77 & 4.44 
\\
     &        &     &       &      &       &        \\
\addlinespace 
Bregman Projections         & 0.05   & No & 2397.21 & 1000   & 2397.21 & 4.43 \\
 &        &     &       &      &        &        \\
\addlinespace 
Bregman Projections & 0.1   & Yes & 115.40 & 187   & 617.11 & 4.40\\
 &        &     &       &      &        &        \\
\addlinespace 
Bregman Projections           & 0.5   & Yes & 19.46 & 252  & 77.22 & 4.59\\
 &        &     &       &      &        &        \\
\bottomrule
\end{tabular}
\end{table}

\newpage

\section{Conclusion}

{ Our new discrete reformulation of the free-support Wasserstein barycenter problem
has a very special structure, which allows for the development of efficient and privacy-preserving federated optimization algorithms.
The main ideas of the reformulation are:
\begin{itemize}
    \item The search for a limited-cardinality barycenter support within a large, but still finite set of candidate points;
    \item The restriction of the barycenter distributions to uniform distributions on their support sets.
\end{itemize}

The idea of the methods is to dualize not only the constraint that limits the support cardinality but also the mass-balance constraints at the support of each individual distribution. As a result of that, the primal variables
representing the barycenter support selection and the couplings can be calculated in a closed form, without any matrix-vector operations or nonlinear functions. Although the couplings may be infeasible, the subgradients with respect to the dual variables are correct and have closed forms as well. This results in a very low iteration complexity.

An essential feature of the proposed method is the protection of data privacy. The coordinator does not have access to the local distributions and receives only aggregate information about the usefulness of the support points for each subproblem.

Our numerical experiments indicate that the dual method is highly efficient and scalable, and the barycenter distributions obtained are accurate.

To the best of our knowledge, our research represents a pioneering effort presenting feasible and scalable algorithms for determining Wasserstein barycenters within the federated setting.}

\section*{Declarations}

This work was supported by the Office of Naval Research award N00014-21-1-2161 and by the Air Force Office of Scientific Research award FA9550-24-1-0284.




\bibliographystyle{abbrv}
 \newcommand{\noop}[1]{}

\end{document}